\documentclass[conference]{IEEEtran}

\makeatletter
\def\ps@IEEEtitlepagestyle{%
  \def\@oddfoot{\mycopyrightnotice}%
  \def\@evenfoot{}%
}
\def\mycopyrightnotice{%
  {\footnotesize XXX-X-XXXX-XXXX-X/XX/\$XX.00~\copyright~20XX IEEE\hfill}% <--- Change here
  \gdef\mycopyrightnotice{}
}

\usepackage{blindtext}
\usepackage{eso-pic}
\IEEEoverridecommandlockouts
\usepackage{cite}
\usepackage{amsmath,amssymb,amsfonts}
\usepackage{algorithmic}
\usepackage{graphicx}
\usepackage{textcomp}
\usepackage{xcolor}
\usepackage{subfigure}
\usepackage{booktabs}
\def\BibTeX{{\rm B\kern-.05em{\sc i\kern-.025em b}\kern-.08em
    T\kern-.1667em\lower.7ex\hbox{E}\kern-.125emX}}
    
\usepackage{eso-pic}
\newcommand\AtPageUpperMyright[1]{\AtPageUpperLeft{%
 \put(\LenToUnit{0.17\paperwidth},\LenToUnit{-2cm}){%
     \parbox{0.9\textwidth}{\raggedleft\fontsize{8}{11}\selectfont #1}}%
 }}%
\newcommand{\conf}[1]{%
\AddToShipoutPictureBG*{%
\AtPageUpperMyright{#1}
}
}

\begin{document}
\title{\vspace*{1cm} Out of Distribution Detection, Generalization, and Robustness Triangle with Maximum Probability Theorem\\
\thanks{This project is supported by the NIH funding: R01-CA246704 and R01-CA240639, and Florida Department of Health (FDOH): 20K04. }
}

\author{\IEEEauthorblockN{Amir Emad Marvasti}
\IEEEauthorblockA{\textit{Currus AI} \\
Orlando, FL \\
amir.emad90@gmail.com}
\and
\IEEEauthorblockN{Ehsan Emad Marvasti}
\IEEEauthorblockA{\textit{Currus AI} \\
Orlando, FL, USA \\
emad@currus.ai}
\and
\IEEEauthorblockN{Ulas Bagci}
\IEEEauthorblockA{\textit{Northwestern University} \\
Chicago, IL, USA \\
ulas.bagci@northwestern.edu}
}

\maketitle
\conf{\textit{Proc. of the International Conference on Electrical, Computer, Communications and Mechatronics Engineering  (ICECCME) \\ 
16-18 November 2022, Maldives}}
\begin{abstract}
Maximum Probability Framework, powered by Maximum Probability Theorem, is a recent theoretical development in artificial intelligence, aiming to formally define probabilistic models, guiding development of objective functions, and regularization of probabilistic models. MPT uses the probability distribution that the models assume on random variables to provide an upper bound on the probability of the model. We apply MPT to challenging out-of-distribution (OOD) detection problems in computer vision by incorporating MPT as a regularization scheme in the training of CNNs and their energy-based variants. We demonstrate the effectiveness of the proposed method on 1080 trained models, with varying hyperparameters, and conclude that the MPT-based regularization strategy stabilizes and improves the generalization and robustness of base models in addition to enhanced OOD performance on CIFAR10, CIFAR100, and MNIST datasets.
\end{abstract}

\begin{IEEEkeywords}
Out of distribution detection, maximum probability theorem, robustness, deep learning, regularization
\end{IEEEkeywords}

\section{Introduction}
\textbf{Regularization.} Training machine learning models requires careful tuning of hyper parameters, architectures, and the loss functions. An effective tuning of the non-trainable aspects of models is mainly justified by empirical evaluations which costs time and is a source of uncertainty in evaluation of models and algorithms. Regularization of models is widely accepted as a way of stabilizing the training process and improving models generalization performance. 

In the Bayesian view of machine learning, the popular regularization approach is to define an explicit prior over parameters of the model. Parallel to this view, a recent approach, called Maximum Probability Theorem (MPT), is introduced \cite{maxprob}. MPT considers a model as an event in the probability space and provides an upper-bound for the probability of the event representing the model. Increasing the probability of a model in MPT leads to regularization effects. Unlike other regularization strategies, MPT eliminates the need for explicitly defined priors over the model's parameters. Instead, MPT requires only a prior on the observables (e.g. input and output) to determine the prior on the parameters or hidden random variables.

\textbf{Our proposal.} We hypothesize that MPT can act as a black-box regularization in vision applications such as OOD detection, and can lead to improved robustness, generalization, and performance in such tasks. We first construct some objective functions encapsulating MPT regularization, and demonstrate their relation with cross entropy loss. Next, we show how to incorporate MPT regularization into training of CNNs and their energy based variants. Finally, we discuss the details of OOD experiments and the results on generalization and robustness.      

\textbf{Why Energy Based Models (EBMs)?} The recent line of work by \cite{zhao2020joint} showed that treating CNNs as EBMs increase the generalization performance of the CNN model and provide additional abilities for the network \cite{JEM} such as:
\begin{itemize}
    \item EBMs can be used to detect anomalous input (Out of Distribution Detection),
    \item EBMs can be trained without labels,
    \item EBMs can be used as generative models.
\end{itemize}
Hence, in this study, we treat CNNs as an EBM \cite{lecun2006tutorial,teh2003energy} and also experiment with effects of MPT on CNN-EBMs. 

The summary of our contributions is below:
\begin{enumerate}
    \item We successfully adapt MPT regularization in the energy based models,
    \item We reconsider the OOD detection problem with the proposed MPT based regularization,
    \item We demonstrate the robustness and generalization effects of MPT regularization on three datasets and 1080 trained networks, with obtained promising results.
\end{enumerate}

\section{Related Work}\label{sec:related}
Our work relies mainly on three lines of research, regularization methods, EBMs, and OOD methods.\\
\textbf{Regularization and Priors.} Regularization and priors of probabilistic models have a rich history, including Jaynes' Maximum Entropy Principle \cite{jaynes1968prior,jaynes1957information,jaynes1957information2}, Jeffreys' uninformative priors \cite{jeffreys1946invariant}, and reference priors \cite{berger2009formal,bernardo1979reference} generalizing Jeffreys priors. Practically, the well accepted approach to regularize probabilistic models is to impose explicit priors on parameters or other variables in the model. In deep learning, the explicit prior is reflected in the popular regularization schemes.
For example, The $\ell_1,\ell_2$ norm regularization reflects a Gaussian and Laplacian prior on variables \cite{krizhevsky2012imagenet,mackay2003information,consonni2018prior}. The logic behind reparameterization invariant priors in \cite{berger2009formal,jeffreys1946invariant}, is that our prior knowledge should not change based on how a parameter is represented or how the model is constructed. The prior on parameters in reference priors, are determined to maximize the mutual information of parameters and the observable random variables \cite{berger1992development,berger1992ordered,berger2009formal,berger2015overall}.
%The work by Jeffreys and related works attempt to pinpoint a prior that is invariant to reparameterization of the variable \cite{jeffreys1946invariant,kass1996selection}.
%The logic behind this form of invariance, is that our prior knowledge should not change based on how a parameter is represented or how the model is constructed.
%Reference priors generalize Jeffreys priors. The prior on parameters are determined by choosing a prior that maximizes the mutual information of parameters and the observable random variables \cite{berger1992development,berger1992ordered,berger2009formal,berger2015overall}. %Reference and Jeffereys priors are determined dependent on the mechanics of models.
Authors in \cite{nalisnick2017variational} propose a gradient based optimization of models with reference priors, which is a promising ground for integrating uninformative priors. However, automatic determination of the reference priors for parameters of complex models is not fully understood \cite{nalisnick2017variational}.

MPT \cite{maxprob}, a recent addition to the probabilistic models literature, offers a  black box regularization of probabilistic models while being invariant to reparameterizations. To use MPT, one does not need to explicitly determine the priors over the parameters. In other words, the parameters do not need to be modeled as random variables. 
%A more comprehensive review of MPT is given in section \ref{sec:MPTover}.

\textbf{Energy Based Models.} EBMs assume some non-normalized log probability distribution over the space of their input \cite{song2021train,lecun2006tutorial}. This non-normalized log probability is referred to as \textit{energy}. In the conventional loss functions such as Log-Likelihood, the loss depends on the imposed log probability distribution. Training EBMs is possible by estimating the gradients and sampling from their imposed probability distribution  \cite{song2021train,hinton2002training,du2021improved}.
The gradients of the log probability with respect to the model's parameters, is in the form of an expectation with respect to the EBMs distribution. If the domain of EBMs are arbitrary large, unbiased and low variance sampling from their underlying distribution is not trivial. To sample from the EBM probability distribution multiple sampling techniques can be used, including the class of MCMC sampling techniques \cite{neal2011mcmc,neal2003slice}. 
The major drawback with the MCMC techniques is that they induce bias in calculation of gradients during their burnout period \cite{mackay1998introduction,propp1996exact}.

Also, depending on the domain of EBMs, the estimated gradients can have large variance \cite{greensmith2004variance}.
The variance in estimation of gradients is a double edge sword. 
On one hand, the variance helps with escaping plateaus in the objective function landscape. On the other hand, large variance in gradients, disrupts convergence of parameters to the optimal point. Despite the bias and variance drawbacks, MCMC methods such as Stochastic Langevin Dynamics \cite{welling2011bayesian} had been shown to be successful in training EBMs \cite{zhao2020joint}.

\textbf{Out of Distribution Detection.} Out of distribution (OOD) detection task is motivated by understanding reliability of prediction of a model on some input data \cite{45512}. Within the existing methods for OOD \cite{hsu2020generalized,mackay2003information,lin2021mood,raghuram2021general} and many others, our work focuses on probabilistic OOD methods and EBMs. We hypothesize that if the model fits a probability distribution to its input, the model's probability distribution function (pdf) on input can be used to identify out-distribution data. Existing OOD detection methods commonly rely on a scoring function that derives statistics from the  output layer of the neural network.
Popular probabilistic measures for OOD are Softmax score \cite{hendrycks2016baseline}, Energy score \cite{JEM}. ODIN \cite{liang2017enhancing}, GODIN\cite{hsu2020generalized} can be viewed as modification of Energy score and Softmax Score.
MOOD \cite{lin2021mood} is an architectural modification using similar probabilistic principles for training and evaluation.

\section{Regularization via Maximum Probability Theorem}\label{sec:MPTover}
\textbf{Notation.} We use the notation according to the MPT paper: a probability space is defined as $(\Omega,\Sigma,P)$ where $\Omega$ is the sample space, $\Sigma$ is the $\sigma$-algebra and $P$ is the probability measure. A parametric model, parameterized by $\theta$ in some arbitrary space, is defined as $M_\theta\in\Sigma$, an event in the $\sigma$-algebra. We use upper case letters for random variables such as $X,Y$ and the range of random variables is denoted by $R(.)$. We use $S(P)$ as a short hand notation to represent support of the probability measure.
We use $x\in S(P)$ to identify outcomes of $X$ with non zero probability. In our construction and similar to MPT, we do not consider $\theta$ as a random variable; only a numerical representation for the event $M_\theta$. For mapping $f$ with a vector space domain and range, the $y$-th entry of the output vector is represented by $f(x)[y]$. 

\textbf{Overview of MPT.} Consider the probability space $(\Omega, \Sigma,P)$, the random variable $Z$ and some event $M_\theta\in \Sigma$, where $\theta$ is a vector in $\Re^t$. MPT shows that the following inequality holds,
\begin{align}
&P(M_\theta)\leq \underset{z\in R(Z)}{\min}\left\{\frac{P(z)}{P(z|M_\theta)}\right\}\label{eq:mpt}.
\end{align}
In MPT, probabilistic models are represented by events such as $M_\theta$, with a known conditional distribution. To regularize the probabilistic models, the upper bound of the model's probability is maximized. The parameter vector $\theta$ is not necessarily considered as a random vector; therefore, explicit definition of a prior on the parameters is not necessary.
As $\theta$ varies, the conditional distribution of the model over $Z$ changes, leading to change in $P(M_\theta)$. Note that the probability upper-bound is equal to 1, only if $P(z)=P(z|M_\theta)$, therefore, $-\log(P(M_\theta))$ can serve as a measure of deviation of the model from the prior.

\textbf{$\alpha$-Parametrization of the MPT.} The \textit{softmin} family of lower bounds for the maximum probability bound is used as a smooth approximation for the $\textrm{min}$ function. The softmin family parameterized by $\alpha>0$ is defined as
\begin{align}
&P_\alpha (M_\theta)\triangleq \left(\sum_{z\in R(Z)}\left(\frac{P(z)}{P(z|M_\theta)}\right)^{-\alpha}\right)^{-\frac{1}{\alpha}}.
\end{align}
The following inequality relates the softmin family and the probability upperbound in (\ref{eq:mpt}):
\begin{align}
&P_\alpha (M_\theta)\leq \underset{z\in R(Z)}{\min}\left\{\frac{P(z)}{P(z|M_\theta)}\right\},
\end{align}
where the equality holds as $\alpha \to +\infty$ :\cite{maxprob}.
The proposed objective function by \cite{maxprob} is $\mathcal{L}(\theta)=P(M_\theta,M^*)$, the probability of intersection of the model and the event $M^*$ representing the true underlying model (the dataset in our case). $M^*$ in our work is modeled with its conditional distribution. The conditional distribution of $M^*$ is the empirical distribution of the dataset. Throughout the paper, $M^*$ is called the \textit{oracle}.

In this paper, we consider the classification problem-OOD detection. We define random variables $X,Y,Z$ to represent the input, label and $Z=(X,Y)$ as shorthand for the concatenation of input and the label. We can write the intersection objective function in its conditional form and treat each term separately,
\begin{align}
\mathcal{L} = \ln P(M^*|M_\theta) + \ln P(M_\theta).
\end{align}
In practice, $M_\theta$ and $M^*$ are implicitly modeled by their conditional distribution over the random variable, $P(X,Y|M_\theta)$. To use the explicit representation, we write the objective function in the marginal form. Prior to deriving the objective function, the relation of $M^*$ and $M_\theta$ needs to be assumed. The choice of modeling the relation of $M^*$ and $M_\theta$ dictates the optimization process. Similar to \cite{maxprob}, the conditional independence of model and oracle is explored here, namely $M_\theta\perp M^*|Z$.

\textbf{Conditional Independence.} If we assume conditional independence between the model and the oracle, we can continue with the marginalization process of the objective function $\mathcal{L}^\perp(\theta)$:
\begin{align}
&\mathcal{L}^\perp(\theta) \triangleq\ln P(M^*|M_\theta) + \ln P(M_\theta),\\
&= \ln \sum_{z\in R(Z)}P(M^*|z,M_\theta)P(z|M_\theta) + \ln P(M_\theta).
\end{align}
Considering the conditional independence assumption, $P(M^*|z,M_\theta)=P(M^*|z)= P(z|M^*)P(M^*)/P(z)$. Assuming that $P(z)$ is uniform, we can treat both $P(z)$ and $P(M^*)$ as constants and drop them for ease of notation. As a result,
\begin{align}
&\mathcal{L}^\perp(\theta)= \ln \sum_{z\in R(Z)}P(z|M^*)P(z|M_\theta) + \ln P(M_\theta).
\end{align}
Finally, to $\alpha$-parametrize $P(M_\theta)$, we can define the following lower bound objective function
\begin{align}
\mathcal{L}^{\perp}_\alpha \triangleq \ln \sum_{z\in R(Z)}P(z|M^*)P(z|M_\theta) + \ln P_\alpha(M_\theta).
\end{align}
Alternatively, because of the concavity of the log function and the Jensen inequality, we can find another lower bound, the cross entropy loss with MPT regularization:
\begin{align}
&\mathcal{L}_\alpha^\perp(\theta)\geq\sum_{z\in R(Z)}P(z|M^*)\ln P(z|M_\theta) + \ln P_\alpha(M_\theta),\\
&\mathcal{L}_\textrm{cross}(\theta)  \triangleq \sum_{z\in R(Z)}P(z|M^*)\ln P(z|M_\theta),
\end{align}
where $\mathcal{L}_\textrm{cross}(\theta)$ is the conventional cross entropy objective function. The summary of the inequality relations is thus
\begin{align}
\mathcal{L}^\perp(\theta) \geq \mathcal{L}_{\alpha}^\perp(\theta) \geq \mathcal{L}_\textrm{cross}(\theta)+\ln P_\alpha(M_\theta).
\end{align}

\section{MPT in Energy Based Models}\label{sec:MPT_EBM}
EBMs add another dimension to the training of classification models. The models are trained on the input distribution without the need for the labels. Here, we incorporate the MPT view of machine learning and the objective functions described in previous section in the formulation of EBMs. %The MPT formulation for the CNN models are included in \cite{maxprob}.

Consider the training set consisted of $N_t$ entries and the test set is consisted of $N_v$ entries, i.e. $D_t=\{z^{(i)}\triangleq(x^{(i)},y^{(i)})\}_{i=1}^{N_t}$.
The training and test set could be understood as an empirical distribution of the underlying true model.
We denote the oracle as $M^*$.
We adapt the intersection objective function described in MPT view.
The classification model parameterized by $\theta \in \Re^t$ is represented by $f(x;\theta)$ where $f:\Re^n \to \Re^m$.
We construct the model's probability distribution $P(x,y|M_\theta)$ using $f$:
\begin{align}
P(x,y|M_\theta) = e^{f(x,\theta)[y]}P(x,y)/\sum_{x',y'\in S(P)} e^{f(x',\theta)[y']}P(x',y').\label{eq:intersect}
\end{align}
Having Eq (\ref{eq:intersect}), and the probability upperbound in (\ref{eq:mpt}), we can write the maximum probability of $M_\theta$ as
\begin{align}
&P(M_\theta) \leq  \frac{\sum_{x',y'\in S(P)} e^{f(x',\theta)[y']}P(x',y')}{\underset{x,y\in S(P)}{\max}\left\{e^{f(x,\theta)[y]}\right\}}.
\end{align}
For $P(M_\theta)$ to be nonzero, either $f$ needs to be bounded or the prior needs to have bounded support. We denote the normalizing constant in (\ref{eq:intersect}) by $\eta(\theta)$.
Under independence assumption, when we add the log-likelihood term, $\eta(\theta)$ will cancel out.
Therefore, in our objective functions only the support of the prior affects the regularization term.
In above equation, we used the prior in the construction of our model.
Without the prior, the probability distribution is not necessarily normalizable \cite{berger1992development}, which leads to instability of the optimization.

\begin{figure*}[t]
    \centering
\subfigure{
\includegraphics[scale=0.36]{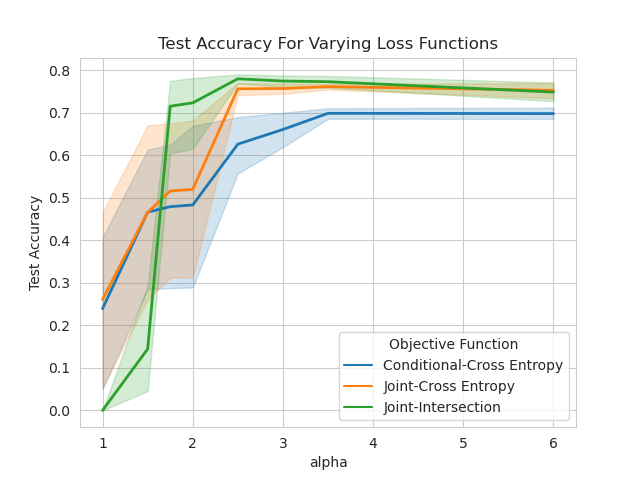}
\label{}}
\subfigure{
\includegraphics[scale=0.36]{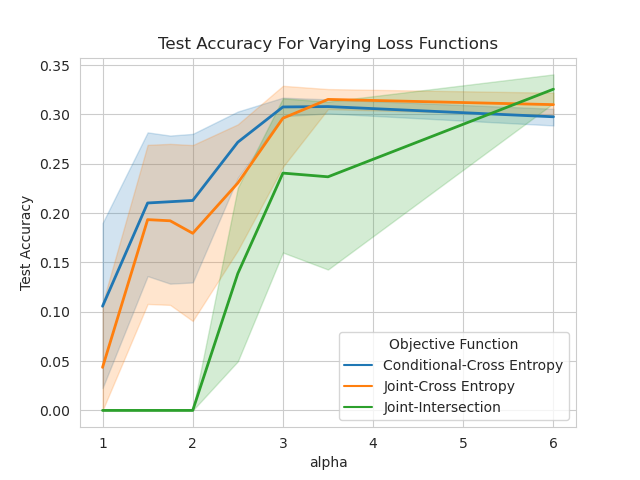}
\label{}}
\subfigure{
\includegraphics[scale=0.35]{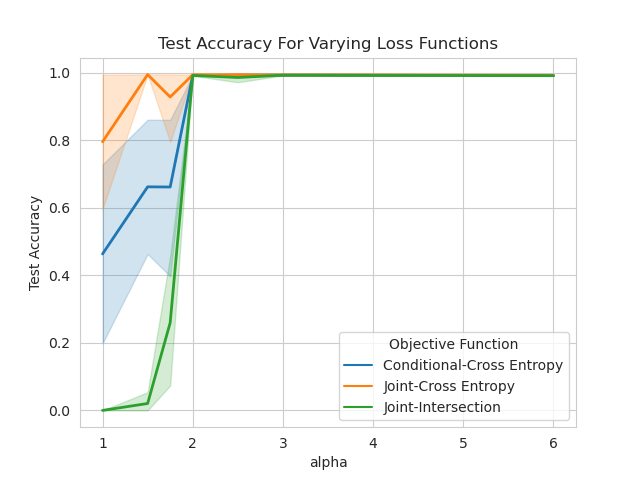}
\label{}}
\caption{The generalization performance of the energy based model with varying hyper parameters and loss functions. Left: CIFAR10, middle: CIFAR100,  right: MINST. }\label{fig:generalizationwhole}
\end{figure*}

\subsection{Conditional Independence Assumption in EBMs}
Having the independence assumption, we can write the intersection objective function for the EBMs as
\begin{align}
&\mathcal{L}_\alpha^\perp(\theta)= \ln \left(\frac{1}{N_t}\sum_{i=1}^{N_t}P(x^{(i)},y^{(i)}|M_\theta)\right) + \ln P_\alpha(M_\theta),\\
& = \ln \left(\frac{1}{N_t}\sum_{i=1}^{N_t}e^{f(x^{(i)},\theta)[y^{(i)}]}P(x^{(i)},y^{(i)})\right) - \ln \eta (\theta) + \ln \eta (\theta)\nonumber\\ &- \frac{1}{\alpha}\ln\left(\sum_{x,y\in S(P)}e^{\alpha f(x,\theta)[y]}\right)\\
& = \ln \left(\frac{1}{N_t}\sum_{i=1}^{N_t}e^{f(x^{(i)},\theta)[y^{(i)}]}P(x^{(i)},y^{(i)})\right) \nonumber\\
&- \frac{1}{\alpha}\ln\left(\sum_{x,y\in S(P)}e^{\alpha f(x,\theta)[y]}\right).\label{eq:finalintersect}
\end{align}

The cross entropy loss can then be written as 
\begin{align}
\mathcal{L}_\alpha^\textrm{cross}(\theta)=& \frac{1}{N_t}\sum_{i=1}^{N_t}\left(f(x^{(i)},\theta)[y^{(i)}]+ \ln P(x^{(i)},y^{(i)})\right),\nonumber\\
&- \frac{1}{\alpha}\ln\left(\sum_{x,y\in S(P)}e^{\alpha f(x,\theta)[y]}\right).\label{eq:finalcross}
\end{align}

The MPT regularization term - the second term in Eq (\ref{eq:finalintersect},\ref{eq:finalcross}) - \textbf{suppresses the probability of the most probable states} in the optimization process.
In a gradient based optimization framework, the difference between cross entropy loss and the intersection loss is that intersection loss re-weights the gradient for each datapoint according to their energy. Whereas, in the cross entropy loss the gradient with respect to the energy of any datapoint is not weighted at all.
%subsubsection{Subset Loss}
% Here we make intersection loss under the subset assumption more clear using the EBM setup
% \begin{align}
% &\mathcal{L}_\alpha^{\subset}= \ln P_\alpha(M^*|M_\theta) + \ln P_\alpha(M_\theta)\\
% &= -\frac{1}{\alpha}\ln \left(\sum_{x,y\in S(P)}\left(\frac{P(x,y|M_\theta)}{P(x,y|M^*)}\right)^{-\alpha}\right)\nonumber\\
% &-\frac{1}{\alpha}\ln \left(\sum_{x,y\in S(P)}\left(\frac{P(x,y)}{P(x,y|M_\theta)}\right)^{-\alpha}\right)
% \end{align}
% If we assume that the data in our dataset is unique we can write
% \begin{align}
% \mathcal{L}_\alpha^{\subset}= -\frac{1}{\alpha}\ln \left(\sum_{i=1}^{N_t}\left(\frac{1}{P(x^{(i)},y^{(i)}|M_\theta)}\right)^{\alpha}\right)\nonumber\\
% -\frac{1}{\alpha}\ln \left(\sum_{x,y\in S(P)}\left(\frac{P(x,y)}{P(x,y|M_\theta)}\right)^{-\alpha}\right) + \ln (N_t)
% \end{align}
% In the subset assumption, the gradients for data points are weighted and the weight is concentrated on the low likelihood data points.

\begin{table}
    \centering
    \setlength{\tabcolsep}{3pt}
        \caption{Classification Results: the empirical mean and standard deviation of training and test accuracy.
    Each statistics were calculated with 120 trained networks with varying hyper parameters.
    The accuracy of diverged models were considered as 0. CCE: Conditional Cross Entropy, JCE: Joint-Cross Entropy, and JI: Joint Intersection.}
\begin{tabular}{llcccccc}
\toprule
      &                    & \multicolumn{3}{c}{Train} & \multicolumn{3}{c}{Test} \\
      &                    &       max &  mean &   std &       max &  mean &   std \\
{} & Objective Functions &           &       &       &           &       &       \\
\toprule
 & CCE &     0.947 & 0.713 & 0.314 &     0.757 & 0.587 & \textbf{0.251} \\
  CIFAR10    & JCE &     0.999 & 0.800 & 0.345 &     0.797 & 0.647 & 0.277 \\
      & JI &     1.000 & 0.847 & 0.335 &     \textbf{0.812} & \textbf{0.666} & 0.264 \\
\midrule
 & CCE &     0.653 & 0.415 & 0.198 &     0.344 & \textbf{0.260} & \textbf{0.111} \\
  CIFAR100    & JCE &     0.616 & 0.376 & 0.220 &     0.372 & 0.245 & 0.135 \\
      & JI &     0.627 & 0.214 & 0.271 &     \textbf{0.397} & 0.134 & 0.170 \\
\midrule
 & CCE &     1.000 & 0.904 & 0.295 &     0.995 & 0.898 & 0.293 \\
 MNIST     & JCE &     1.000 & 0.990 & 0.098 &     \textbf{0.996} & \textbf{0.985} & \textbf{0.097} \\
      & JI &     1.000 & 0.753 & 0.426 &     0.995 & 0.748 & 0.423 \\
\bottomrule
\end{tabular}

    \label{tab:generalization}
\end{table}
\begin{figure*}[t]
    \centering
\subfigure{
\includegraphics[scale=0.35]{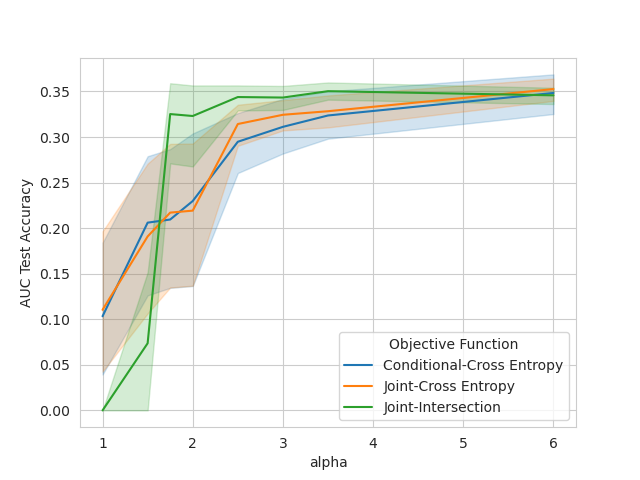}
\label{fig:rob1}}
\subfigure{
\includegraphics[scale=0.35]{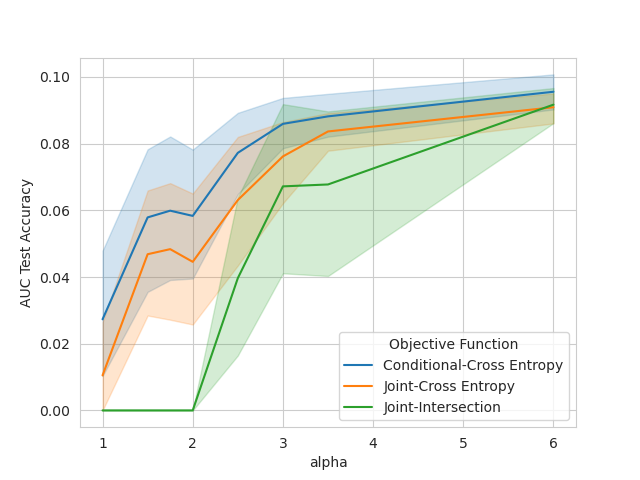}
\label{fig:rob2}}
\subfigure{
\includegraphics[scale=0.35]{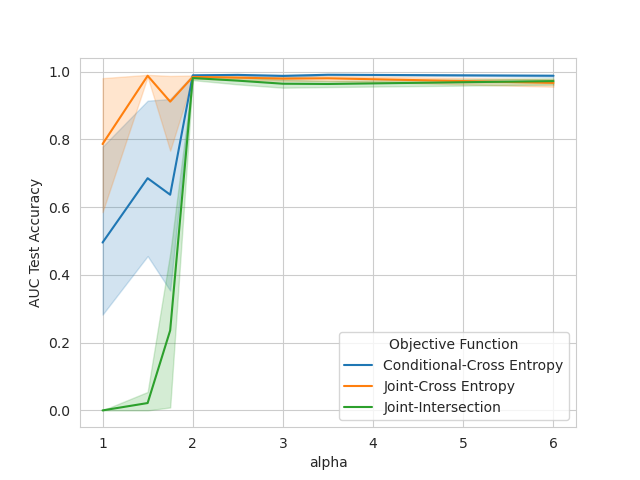}
\label{fig:rob3}}
\caption{The robustness of the energy based models with varying hyper parameters and loss functions. The experiment shows how the trained models are robust to additive noise to the input. The vertical axis represents the mean of the test accuracy of the trained networks for varying noise levels. }\label{fig:robust}
\end{figure*}

\subsection{Approximating the gradients}
To have an unbiased approximation of the gradients in the intersection objective function, we need to use MCMC methods. Drawing unbiased approximation of gradients using MCMC methods requires passing the burnout period. Approximating the burnout period is an open problem in the MCMC literature. We accept the bias in the gradients for the benefit of computational efficiency.

In the intersection objective function, the total gradient is not a linear function of sub-gradients corresponding to the subsets of data. The gradient cannot be approximated with small batches without bias. Considering limit cases, the nature of the bias in our approximation becomes clear. If we choose a batch size of one, the sub-gradients are unbiased approximators of the cross entropy's total gradient, and as the batch size increases, the sub-gradients approximates the intersection objective function' total gradient. Based on this reasoning, we accept the trade off and allow the sub-gradients to be biased in the favor of efficiency.

We can use a small batch size and approximate the loss function.
%Biased Approximation of the gradients with small batches, is similar to importance sampling with a small batch size.
When the loss function is approximated we can calculate the gradients using any automatic gradient calculation framework.
\begin{table}[h]\small
\setlength{\tabcolsep}{12pt}
    \centering
    \caption{
    Robustness of the trained models to noise level. The AUC score is the average of the test accuracy for varying uniform noise levels, added to the inputs. CCE: Conditional Cross Entropy, JCE: Joint-Cross Entropy, and JI: Joint Intersection.}
    \label{tab:robust}
\begin{tabular}{lccc}
\toprule
{} & \multicolumn{1}{c}{CIFAR10} & \multicolumn{1}{c}{CIFAR100} & \multicolumn{1}{c}{MNIST} \\
\midrule
CCE   &    0.34$\pm$ 0.05 &     0.09 $\pm$ 0.01 &  0.99 $\pm$ 0.00 \\
JCE         &    0.34 $\pm$ 0.03 &     0.09 $\pm$ 0.01 &  0.97 $\pm$ 0.02 \\
JI &    0.35 $\pm$ 0.02 &     0.08 $\pm$ 0.04 &  0.97 $\pm$ 0.02 \\
\bottomrule
\end{tabular}
\end{table}

\subsection{Max Energy Score and OOD Detection}
Energy based model fits a distribution to the input domain as well as the labels.
The energy score (i.e., likelihood of observations of some input data up to a normalization constant) can be used to discriminate between data that is observed in training and the data that was not observed.
In addition to energy score and softmax score \cite{hendrycks2016baseline}, we introduce a new score for discriminating between in and out distribution data; max energy score: 
\begin{align}
    S_\textrm{ME}(x) = \underset{y\in S(P)}{\max}f(x;\theta)[y'].
\end{align}
This new scoring value (max energy score) is simply defined as the maximum energy of the labels per input. Smaller the score is, the more highly that input data belongs to out-distribution.

\section{Experiments and Results}\label{sec:Exp}
\begin{figure*}[t]
\subfigure{
\includegraphics[scale=0.5]{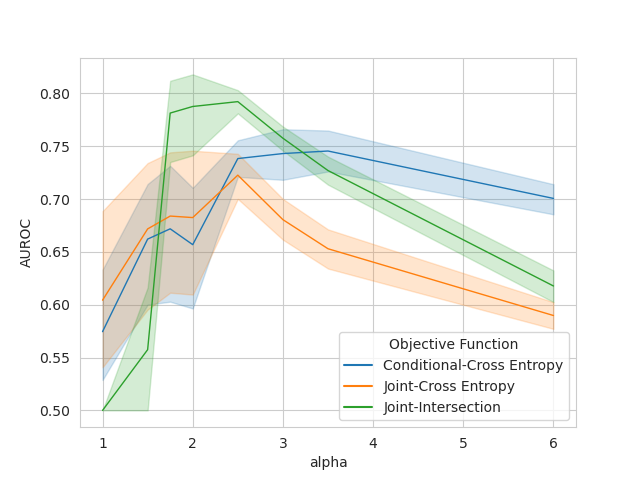}
\label{fig:ood:energy}}
\subfigure{
\includegraphics[scale=0.5]{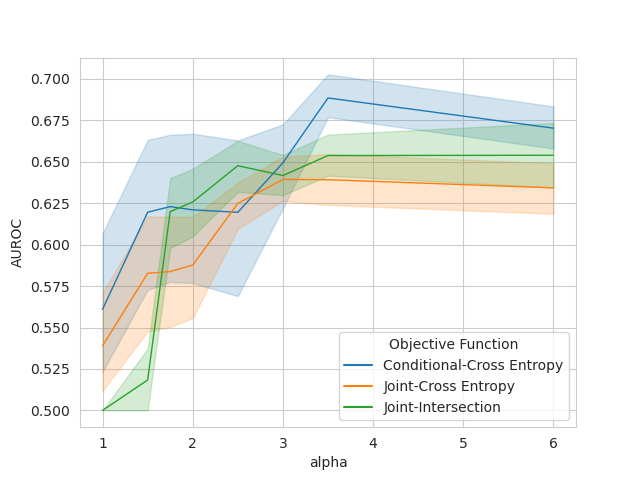}
\label{fig:ood:softmaxscore}}
\subfigure{
\includegraphics[scale=0.5]{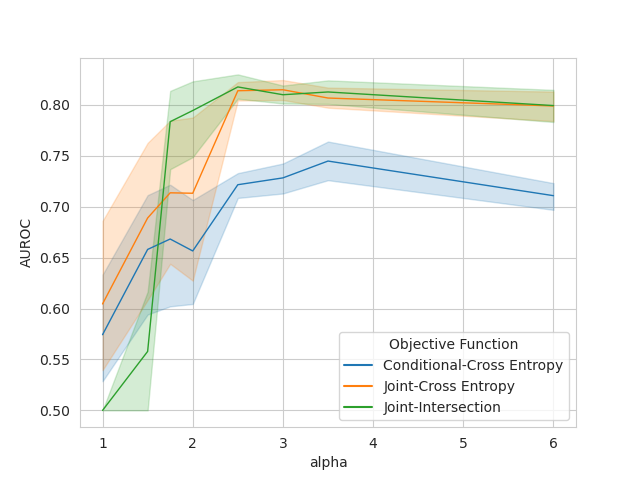}
\label{fig:ood:maxenergy}}
\subfigure{
\includegraphics[scale=0.5]{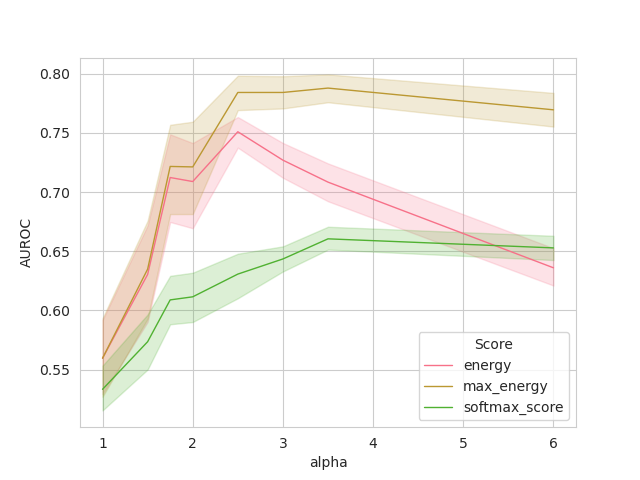}
\label{fig:ood:comparison}}
\caption{
The AUROC of the trained networks per value of $\alpha$. The in dataset was chosen to be CIFAR10. The AUROC was averaged over 6 out datasets. Variations of the AUROC versus changes of the $\alpha$ hyperparameter in MPT for different objective functions are shown. Upper left: Energy Score, upper right: Softmax Score, bottom left: Max Energy, bottom right: Score Comparison.}\label{fig:ood}
\end{figure*}
We used a 12 layer CNN to train on CIFAR10, CIFAR100 and MNIST datasets. We chose to use CReLU activation \cite{Crelu} instead of the ReLU because of its gradient norm preserving property.
For all convolution layers we used 64 filters. The CNN does not have Batch Normalization~\cite{ioffe2015batch} and Dropout~ \cite{hinton2012improving}. We avoided any other probabilistic layer to avoid their interference with the optimization.
Including probabilistic layers in our experiments might interact with the objective function we proposed.
Their effect is out of the scope of the current study.
The purpose of using the described model was to experiment with a backbone and simple architecture.
We avoided hyperparameter tuning as much as possible and tested all the variants with the same hyperparameter range. The learning rate was fixed to 0.01. 
We experimented with different batch sizes to include the effect of the bias of subgradients in our experiment.
The batch sizes in our experiments were chosen from 64, 128, 256. Each hyperparameter setup was trained 5 times.
The total number of networks trained are 360 for each dataset.

The experiment is designed to show the effects of using MPT with the proposed objective functions on the CNN and its energy based variant on the classification task.
In MPT based objective functions, the $\alpha$ hyper parameter determines how much the model is regularized. 
\textbf{A comparative study is embedded in our experiments.}
Setting $\alpha$ to 1, removes the regularization effect of MPT and is equivalent to the usual likelihood training methods.
We have included $\alpha=1$ to compare MPT with the conventional cross entropy and energy based loss.

Our experiments are three fold, (i) we experiment with classification task to show the stability of training and generalization performance having the MPT regularization.
(ii) we experiment the robustness of trained networks to the additive noise to input. (iii) We demonstrate the effect of MPT regularization on OOD task. We use the the popular metric, Area Under the Receiver Operating Characteristics (AUROC) to evaluate the OOD performance.

\begin{table}
    \centering
    \setlength{\tabcolsep}{5pt}
%\resizebox{1\textwidth}{!}{
% \begin{minipage}{\textwidth}
\caption{OOD AUROC scores. The empirical mean and standard deviation of the used scores for discrimination. Each statistics were calculated with 120 trained networks with varying hyper parameters. The AUROC of diverged models were considered as 0.         \label{tab:OOD}}
\begin{tabular}{lccc}
\toprule
{Objective Function} & \multicolumn{3}{c}{Mean AUROC} \\
\cmidrule(lr){2-4}
{} &  Energy & Max Energy (Ours) & Softmax Score  \\
\midrule

Conditional Cross   &   \textbf{0.687} $\pm$ 0.102 &      0.683 $\pm$ 0.098 &         0.632 $\pm$ 0.084  \\
Joint Cross         &   0.661 $\pm$ 0.110 &     \textbf{ 0.744} $\pm$ 0.130 &         0.604 $\pm$  0.060  \\
Joint Intersection &   0.690 $\pm$ 0.124 &      \textbf{0.734} $\pm$ 0.135 &         0.608 $\pm$  0.067 \\

 & \multicolumn{3}{c}{Max AUROC} \\
\cmidrule(lr){2-4}
{} &  Energy & Max Energy (Ours) & Softmax Score  \\
\midrule

Conditional Cross   &     0.824 &      0.819 &         0.743 \\
Joint Cross         &     0.836 &     \textbf{ 0.854} &         0.713 \\
Joint Intersection &     0.853 &     \textbf{ 0.866} &         0.707 \\
\bottomrule
\end{tabular}

% \end{minipage}

\end{table}

\subsection{Training stability and Generalization performance}
The results of the generalization performance of the trained networks are demonstrated in Fig \ref{fig:generalizationwhole}. The summary of the performances can be found in Table \ref{tab:generalization}. 
The loss functions used are Conditional Cross Entropy, Joint Cross Entropy and Joint Intersection.
Conditional Cross Entropy is the usual cross entropy loss function.
Joint Cross Entropy is the energy based variant with the cross entropy loss. Joint Cross Entropy enforces the network to train jointly on the input and the label domain. 
Joint Intersection is the energy based loss function derived from MPT described in equation (\ref{eq:finalintersect}).
All of the loss functions were equipped with MPT regularization, where the effect of regularization is controlled by $\alpha$.

Figure \ref{fig:generalizationwhole} shows that increasing the $\alpha$ parameter has positive effect on both generalization and stability of training.
On all datasets, as $\alpha$ increases the average of test accuracy of the models increases and the empirical variance decreases. 
During training phase some networks diverged. The accuracy of the diverged networks were set to 0, impacting the average accuracy.
The best performance in our experiments is subpar to the state of the art results on CIFAR datasets.
The inferiority of our results could corresponds to almost blind tuning of hyperparameters and the network architecture.
Our experiment represent the practical situation were the dataset at hand is new and the initial attempts for hyperparameter tuning is suboptimal. 

Table \ref{tab:generalization} shows the max and average test accuracy of the trained networks; the $\alpha$ parameter settings were marginalized out. The test accuracy on CIFAR100 is exceptionally low. The low test accuracy can be explained with the low average training accuracy.
The low test accuracy can potentially be solved by increasing the capacity of the network.
The maximum test accuracy, in CIFAR10 and CIFAR100, was achieved by Joint-Intersection objective function.
In MNIST the difference of the intersection test accuracy and the cross entropy is insignificant.
The results show that cross entropy is a safe choice for training under-regularized loss functions.
But with the heavier regularization, the intersection loss is superior.

\subsection{Robustness Analysis} 
We compare the performance of the trained models, when the input data is perturbed with some level of noise. For each network, we add uniform noise to the input data, and test the performance of network for different levels of noise. The prior for the input in all our experiments is a uniform distribution with support of $[-1,1]$. We clamped values, after adding the noise, so that the noisy input falls in the support of the prior. We evaluated the test accuracy of each network by measuring the test accuracy per noise level. The support of the uniform noise were selected from [0.1,0.2,...1] times the support range of the prior. 
For each network we approximated the Area Under the Curve (AUC) of the noise level and the test accuracy.
Figure \ref{fig:robust} demonstrates the variation of the AUC measure with respect to changes in $\alpha$. 
The results are consistent with respect to the results of Figure \ref{fig:generalizationwhole}. As $\alpha$ increases the models are more robust to changes in the input distribution.
Table \ref{tab:robust} compares each loss function's AUC performance by averaging $\alpha$ out. Without considering the regularization, the results indicate that the objective functions, on average, behave similarly. However, Joint-Intersection responds to the regularization better than the rest as seen in Fig \ref{fig:robust}. %To calculate a final measure of robustness we estimated the area under the curve for the networks.
\begin{table}[]
    \setlength{\tabcolsep}{5pt}
\caption{Comparison of the on OOD AUROC scores reported in reference \cite{lin2021mood} with our best result. Except the last row (where we used 6 datasets), all other rows include 10 datasets.}
    \centering
    
    \begin{tabular}{l c c c }
    \toprule
         & Score & Model (\#layers) & AUROC\\
        \cmidrule{1-4}
        MOOD\cite{lin2021mood,huang2018multiscale} & - & MSDNeT (20) & \textbf{0.912}  \\
        ODIN\cite{liang2017enhancing,wideresnet} & Softmax& WideResNet(40)& 0.901\\
        Mahalonobis\cite{mahalonobisOOD} & -&WideResNet(40) &0.893 \\
        Mahalonobis\cite{mahalonobisOOD} & -&MSDNeT (20) &0.828 \\
        Liu et al.\cite{energy_based} & Energy& WideResNet (40)&0.900\\
        Liu et al.\cite{energy_based} & Energy& MSDNet(20) (40)&0.904\\
        EBM+MPT (Ours) & Max Energy& CNN (12) (40)&0.866\\
        \bottomrule
        
    \end{tabular}\label{tab:compareall}
    
    \label{tab:my_label}
\end{table}

\subsection{OOD Detection}
We use the trained network and test the out of distribution detection performance of the networks on 6 datasets.
The setup of our evaluation is similar to that of \cite{lin2021mood}. The in-distribution dataset in our experiments is CIFAR10 \cite{CIFARs}, and the out distributions are MNIST \cite{mnistdata}, KMNIST \cite{kmnist}, SVHN \cite{SVHN}, CIFAR100 \cite{CIFARs}, STL10 \cite{STL10} and Fashion MNIST \cite{fashionmnist}. We use the softmax score, energy score, and the proposed max-energy score to discriminate between in and out distribution. We use the data from the test set of the in-distribution and we used both train and test data for the out-distribution. We used conventional AUROC score for evaluation of OOD strategies \cite{lin2021mood,hsu2020generalized,hendrycks2016baseline,liang2017enhancing}.
The AUROC is averaged over the out-distribution datasets.

The average AUROC of trained networks with respect to the $\alpha$ parameter is shown in Fig \ref{fig:ood}. %As opposed to previous experiments, there is a sweet spot for $\alpha$ parameter in the OOD task: $\alpha=2$ has a special property that the optimal solution for the model distribution matches the oracle distribution. In our experiments the sweet spot is not necessarily $\alpha=2$, since our gradients are biased. 
Decrease of the AUROC for large values of $\alpha$ is expected. As a reminder, energy based models assume a probability distribution on the input domain. As $\alpha$ increases further form 2 (see reference \cite{maxprob} for significance of $\alpha=2$), the optimal probability distribution solution tends to be closer to the prior distribution. This means that out-distribution scores would be similar to the in-distribution scores. The Joint-Intersection objective function is superior for both Energy score and the Max Energy score. For Softmax Score, the conditional cross entropy is the superior objective function.

Fig \ref{fig:ood:comparison} represents the comparison of different energy scores with respect to $\alpha$. As seen Max Energy score performs better in the OOD task compared to the other OOD scores. Table \ref{tab:OOD} summarizes the results for the energy scores and the loss functions.
Table \ref{tab:compareall} compares the results of different methods on OOD, obtained by reference \cite{lin2021mood} to provide a sense of performance scale for the reader. 
The inferiority of our results could be attributed to the choice of a light weight and vanilla CNN model with only 12 layers.
MOOD \cite{lin2021mood}, ODIN\cite{liang2017enhancing}, JEM\cite{JEM,energy_based} and Liu et al.\cite{energy_based}, use modified versions of EBM-CNNs and the OOD scores discussed in our paper. 
In Figure \ref{fig:ood}, We have shown that MPT regularization consistently improves the OOD performance of the popular OOD scores and the performance of EBM-CNNs. 
Therefore MPT has the potential for integration into the mentioned baselines.

\section{Discussion and Future Research}\label{sec:discussion}
%The experiments were designed to show the effects of using MPT and the loss function variants with respect to hyperparameter tuning.
The empirical aspect of our research is mainly ablation studies and proof of concept for effectiveness of MPT.
Our empirical evidence holds, at least, when the model architecture and the hyper parameters are not tuned to the dataset.
It is hard to confidently assert that MPT works for the vast variations of models that currently exist in the machine learning scope. However, so far we have not identified any incompatible probabilistic model in our empirical evidence.
Furthermore, since the proof for MPT does not rely on any assumptions, it is reasonable to expect that the framework is general enough.

Given that MPT is theoretically well grounded, and offers a black box view on priors, it is a promising direction for further investigation.
Also, the empirical evidence, although naturally limited, is inline with what is expected from theory. 
For example, the generalization of the model using MPT can be tied to smoothness of model's distribution on input/label, $P(x,y|M_\theta)$.
The test accuracy of a model is only dependant on the test data distribution and the model's distribution.
Maximizing the model probability forces the model's distribution to be close to the prior distribution (e.g. Uniform).
If the prior is smooth, maximizing the probability of the model enforces the model's distribution to be smoother.
Models assuming smoother conditional distributions are more invariant to changes of input/label distribution. 
Therefore, the smoother models are more robust to the difference of the train and test distribution.
We intend to follow up on this thought for future research and verify whether probability of the models could be tied theoretically to their generalization performance.

MPT does not restrict the possibility of having explicit priors on parameters. MPT provides an upper bound for the prior density of parameters. Any explicit assumption made about the parameters can be combined with MPT to form a new prior. For example including the explicit prior $q(\theta)$ can be done by constructing the density, $p(\theta) = P_\alpha(M_\theta)q(\theta)$ followed by normalization. So far the benefits and adverse effects of combining explicit priors is not clear to us. 
It is possible to speculate that the gradients of regularization may be affected by the gradient vanishing problem in the MPT framework. Having explicit priors could help with the gradient vanishing of regularization.
We intend to test the effect of incorporating explicit priors to the MPT framework in our future experiments.

\section{Conclusion}
In this paper, we have incorporated Maximum Probability Theorem  into training of Energy Based Models (EBMs) and demonstrated its black-box regularization property in classification and out of distribution detection problems. Obtained from six publicly available datasets, our experiments demonstrated that (1) learning input and the label jointly in EBMs with adequate regularization outperform CNNs with softmax layer. (2) MPT regularization, without any exception in our experiments, increases the stability of training, lowers the variance of test accuracy, and improves generalization performance. (3) Incorporating MPT regularization improves AUROC in the OOD tasks.

\bibliographystyle{splncs04}
\bibliography{energympt}

\end{document}